\pdfoutput=1

\documentclass[12pt,journal,compsoc]{IEEEtran}

\usepackage{graphicx}
\usepackage{multicol}
\usepackage{amsmath,amssymb}
\usepackage[table]{xcolor}

\usepackage{adjustbox}
\usepackage{subcaption}
\usepackage{slashbox}
\usepackage[switch]{lineno} 
\usepackage{lipsum} 

\ifCLASSOPTIONcompsoc
\else
\fi
\ifCLASSINFOpdf
\else
\fi

\begin{document}
%

\title{Gaze Distribution Analysis and Saliency Prediction Across Age Groups}
\author{Onkar Krishna\textsuperscript{1}, Kiyoharu Aizawa\textsuperscript{1}, Andrea Helo\textsuperscript{2}, R\"{a}m\"{a} Pia\textsuperscript{2}; Dept. of Information and Communication Engineering, The University of Tokyo, Japan\textsuperscript{1};  Laboratoire Psychologie de la Perception, Universit\'{e}  Paris Descartes, France\textsuperscript{2}}

\IEEEtitleabstractindextext{%
\begin{abstract}
Knowledge of the human visual system helps to develop better computational models of visual attention. State-of-the-art models have been developed to mimic the visual attention system of young adults that, however, largely ignore the variations that occur with age. In this paper, we investigated how visual scene processing changes with age and we propose an age-adapted framework that helps to develop a computational model that can predict saliency across different age groups. Our analysis uncovers how the explorativeness of an observer varies with age, how well saliency maps of an age group agree with fixation points of observers from the same or different age groups, and how age influences the center bias. We analyzed the eye movement behavior of 82 observers belonging to four age groups while they explored visual scenes. Explorativeness was quantified in terms of the entropy of a saliency map, and area under the curve (AUC) metrics was used to quantify the agreement analysis and the center bias. These results were used to develop age adapted saliency models. Our results suggest that the proposed age-adapted saliency model outperforms existing saliency models in predicting the regions of interest across age groups.
\end{abstract}

}

\maketitle
\IEEEdisplaynontitleabstractindextext
\IEEEpeerreviewmaketitle

\section{Introduction}
Computational models of human visual attention are becoming increasingly important, and investigations of these have driven much research by psychologists, neurobiologists and researchers in computer vision. The problem of predicting a region of a scene that attracts the observer remains a core challenge in vision research that can at present be solved in two ways: using eye-tracking devices, like the TobiiX50 \cite{a1}  and Eyelink1000 \cite{a2} and, by developing a computational model \cite{a3,a4,l,n} to mimic human vision for scene-viewing. Although eye trackers achieve high prediction accuracy, they are not always an in-hand option \cite{n}. Thus, the use of computational models has gained an importance in the last few decades. 

   The era of the development of computational models was heralded by the pioneering work of Itti et al. \cite{l} based on Treisman's feature integration theory (FIT) \cite{j}, where a master saliency map is obtained by combining bottom-up feature maps in parallel. A series of works \cite{faa}, \cite{fab}, \cite{faba}, \cite{fabab}, \cite{fababa,mtmt} have since investigated similar issues, where the major differences lay in the way the features were selected and maps combined. Some researchers integrated the maps linearly whereas others used non-linear techniques to combine them \cite{fabab}, \cite{fababa}. The next set of saliency models \cite{a5}, \cite{a6} were based on top-down factors, which are, the given task  \cite{f}, human tendency \cite{bb}, habituation and conditioning \cite{c}, and emotions \cite{d} as these factors are closely related to visual attention during scene viewing.
   
 Even though eye-movement control improves extensively already during early infancy, an adult-like control is reached later during childhood \cite{a7}. For example, the capability to fixate a target is acquired during the first few months of life \cite{a71,a72} but more complex aspects of the fixation system, such as steadiness of fixations and cognitive control continues to develop until adolescence \cite{a7}. Studies on development of saccade control found that the saccades were shorter and less precise when comparing children with adults \cite{a7}. Furthermore, cognitive control of saccade execution, operationalized by the performance in pro and anti-saccades tasks, reaches an adult-like performance level at around 10 to 12 years of age \cite{a73}, \cite{a8}, \cite{a81}.  
   
 Supporting evidence from developmental studies on scene exploration \cite{h}, \cite{i} has shown that there are remarkable differences in the scene-viewing behavior of observers across age groups. For example, local image features, such as color, intensity, luminance, etc., were shown to guide fixation landings more early in childhood, while later in childhood, fixation landings are more dominated by top-down processing \cite{h}, \cite{i}. 
  
  In spite of a few studies reporting developmental changes in scene viewing behavior, there are no studies that have systemically analyzed the gaze allocation of observers across age groups using computational models. So far, computational models have relied on the data collected in adult participants but due to significant changes in visual skills during the development, it is essential to include also age factor to the computational models 
  
  Thus, computational models that have been developed until now compromise on prediction accuracy as they do not take into account age factors. Our study aims to develop a new computational model that includes observers age in predicting salient locations for an image. Our study is divided into two part: the first part consists of quantitative analysis of the age-related differences in fixation landings during scene viewing, and the second part consists of, our proposed age-adapted computational model of saliency prediction based on the analysis results reported in the first part. The framework of the proposed study is reported in following section and shown in Fig. 1.  
  
  \begin{figure*}[!h]
  \centering
 \includegraphics[width=17cm, keepaspectratio]{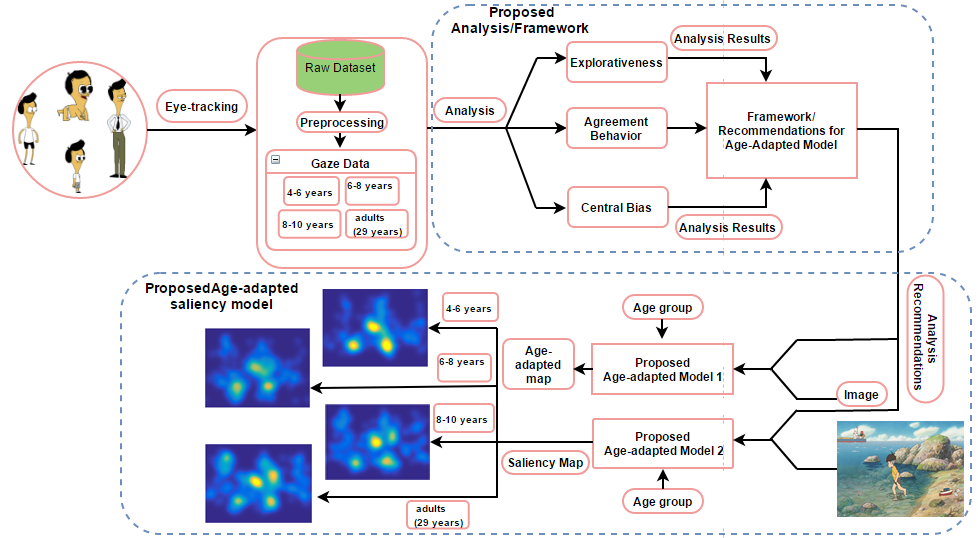}%
\caption{{\bf The framework of our proposed study:}
 It consist of two parts, analysis part and proposed age-adapted saliency model as shown in figure, 29 years is mean age of adult observers}
\label{fig1}
\end{figure*}

\subsection{Framework of our study}

We focused on understanding age-related changes in scene-viewing behavior and developing formal measures to quantify these differences as shown in proposed analysis part of Fig. 1. The next part of the study was to develop an age-adapted saliency model that incorporates these changes and reflects age-related differences in predicting the saliency map. The flow diagram of the proposed age-adapted saliency model is in shown Fig. 1. Our work is strategically beneficial, as most conventional models of visual attention can be easily tuned to age-related changes in observers by following the recommendations of the results of our analysis. The following framework is used in the proposed study.\\

\textbf{Analysis} 
We selected fixation landing locations as a main attribute to analyze the age-related differences in scene viewing behavior. The reason for this selection relies on the fact that the purpose of our analysis was to develop an age-adapted saliency model and the existing saliency models consider fixation location as a key gaze attribute in predicting salient regions. The analysis was mainly focused on three aspects of scene viewing behavior: explorativeness, agreement within and between age groups, and center bias. This selection was based on the fact that the previous studies \cite{n}, \cite{faba} have analyzed the fixation spread and center bias tendency in order to propose a better model of saliency prediction. Similarly, our analysis results of age-related differences in fixation distribution can assist in developing an age-adapted saliency model. The explorativeness index quantitatively measures the spread of fixation locations.

   The second metrics, called ``agreement score", indicates how well the observers within same age group or of different age groups agrees in terms of explored locations. The center bias was used to revel the age-related differences in center bias tendency. \\

\textbf{Age-adapted saliency model}
Based on our analysis, we proposed an age-adapted framework i.e. analysis recommendations, which can be used to upgrade available saliency models. This new age-adapted framework simulates the elements of a visual scene that are likely to attract the gaze of observers across age groups. As discussed earlier, most saliency models use Feature integration theory, where weighted combinations of feature maps of all scales are calculated to determine salient locations. Instead, we chose selectively the scales for different age groups depending on the level of details they observed.

\subsection{Existing Saliency Models}

The human vision system has been studied extensively, and several theories have been proposed to explain how our visual system process information. Feature integration theory (FIT) \cite{j} is one of the most important psychological theories used to develop visual attention models. It suggests that the set of features for a given visual scene is processed automatically and in parallel during early stages of viewing to obtain conspicuous locations. These features are combined in late phase of viewing to help in object identification and separation. 

It was subsequently found that human visual behavior is not only affected by scene-related bottom-up features, but also by top-down features. A guided search model \cite{k} was proposed to account for the influence both of bottom-up and top-down features on human visual behavior. 

In the last decade, many of image processing and computer vision researchers have used these theories \cite{j}, \cite{k} to make computers mimic our visual system, however, all these models are developed for young adults.  We briefly review some of these models according to the techniques and/or features they use.\\

\textbf{Bottom-up features based models}    
Itti et al.'s model \cite{l}, implemented over FIT theory is one of the most well- known models, where bottom-up features of a scene are extracted in parallel by a set of linear center-surrounded operations similar to the visual receptive field. The normalized values of the extracted features are then fused in the later stages to obtain conspicuous locations of an image.

The graph-based visual saliency model \cite{a4} also follows a similar approach in generating the activation maps of different feature channels at multiple spatial scales. Furthermore, these maps are represented as fully connected graph, where the equilibrium distribution in a Markov chain is treated as the saliency map. However, these models extract features over a fixed spatial scale, and the age-related changes in image feature-related viewing \cite{h} were not considered while generating a master saliency map.

\textbf{Combination of bottom-up and top-down features}
Another approach to developing visual saliency models involve combining low-level cues, i.e., bottom-up information and top-down factors to generate the conspicuity map. Torralba (2003) \cite{faa} and Torralba (2006) \cite{fab} proposed a model using a Bayesian framework that integrates the scene context with a bottom-up saliency map. 

Similar to the Bayesian framework, the SUN model of saliency prediction \cite{faba} combines bottom-up features represented as self-information with top-down information, where top-down information is represented either by Difference of Gaussian (DoG) or independent component analysis (ICA) features extracted from images. Some studies have integrated scene related factors with the human tendency for top-down cues, such as face, objects detectors, and the center bias. A boolean map based model \cite{m} was recently developed based on Gestalt psychological studies \cite{bb}, and outperformed other state-of-the-art models on saliency related datasets. However, these models do not take into account developmental studies reporting that bottom-up processing is dominanting during early development while the ifluences of top-down processing increase with increasing age \cite{h}, \cite{a9}, \cite{a10}, \cite{a11}.\\

\textbf{Patch based models} 
Patch based dissimilarity measures are another line of approach where saliency is estimated in terms of dissimilarity among neighbouring patches. A patch-based saliency estimation method \cite{fabab} was proposed to compute saliency using dissimilarity among patches. This was measured by the average distance of regional covariance among neighbouring patches. First-order image statistics such as difference of mean value is also incorporated with this algorithm to obtain better results. 

Another patch based method\cite{fababa} was proposed to estimate the saliency of each patches by measuring the spatially-weighted dissimilarity among them, where the image patches were represented in reduced dimensional space by applying principal component analysis (PCA). These models are not suitable for age-adapted prediction of salient locations as the optimal patch size is selected for the highest prediction accuracy over the eyetracking data collected for young adults only.\\

\textbf{Models based on Supervised Learning on Eye tracking datasets}  
Supervised learning-based methods using eye-tracking data collected from young adults constitute another technique to build computational models. \cite{n} Proposed a model that simply learns to predict saliency from an eye-tracking dataset containing over 1003 images viewed by 15 young adults. 

Some of the eye-tracking datasets used for these learning methods are listed in Table 1. It can be seen from the table that the participants of these eye-tracking experiments across all datasets were adults (aged 18 to 45 years). Thus all state-of-the-art models to predict visual saliency using these datasets are inclined to reflect the scene exploration behavior of adult observers only.
 
 \begin{table}[]
    \center
    \caption{Saliency benchmark dataset}
   \begin{tabular}{l*{6}{c}r}
   
 Dataset & Images & Observers & Age & Duration(s)  \\
\hline
MIT300\cite{a12} & 300 & 39 & 18-50 & 3  \\
FiWI\cite{a13} & 149 & 11 & 21-25 & 5 \\
NUSEF\cite{a14}& 758 & 25 & 18-35 & 5  \\
DOVES\cite{a15}& 101 & 29 & 27 & 5 \\
Toronto\cite{a16}& 120 & 20 & 18-22 & 4 \\
\end{tabular}
\end{table} 

\subsection{Eye tracking data}
  
  \textbf{Subjects and stimuli}  We analyzed the eye-tracking dataset collected in \cite{i}. The eye-tracking data was obtained for 82 observers from different age groups. All observers had normal or corrected-to-normal vision. Participants were assigned to 4 different groups: four-six years, six-eight years, eight-ten years, and adults (mean age, 29 years). We use 4 year, 6 year, 8 year, and adults to refer these groups in order. The experiment was conducted on images of $1024\times764$ pixels. The images were taken from children’s books and movies, and characterized to have eventful backgrounds.  \newline 
  
\noindent \textbf{Apparatus and Procedure} The remote eye-tracking system EyeLink 1000 with a sampling rate of 500 Hz was used to measure eye gaze, and provided us with the raw data that was sampled to obtain fixations and saccades. The spatial resolution of eye tracker was below $0.01^{\circ}$, and spatial accuracy more than            $0.5^{\circ}$. The random fixations and noise were discarded by processing the raw data by fixation detection algorithm supplied by SR research (EyeLink).

The following procedures followed during eyetracking experiment: 
\begin{enumerate}
\item A five point calibration and validation was performed before starting the experiment, and subjects were asked to explore the scene which was presented for 10 seconds. 
\item Further the scene was subsequently replaced by an image segment and participants had to determine if the segment was part of previous scene or not. The segment recognition test was included to maintain motivation of our participants and also to understand the levels of engagement i.e. how engaged the participants were in the material. The results of the task performance reported in \cite{i} suggests the high level of engagement for the selected stimuli for all age groups, which also confirms the age appropriateness of our selected stimuli.  
\item Picture were viewed at a distance of 60 cm from a screen at a resolution of $1024\times728$. 
\end{enumerate}
      
 \noindent\textbf{Data Representation} For each image, fixation landings of all observers were used to generate two maps: a human fixation map and a human saliency map. The human fixation map was a binary representation of fixation locations, and the human saliency map was obtained by convolving a Gaussian filter across the fixation locations, as in \cite{n}. The visualizations of human fixation and human saliency maps are shown in Fig. 2. These maps were used to analyze eye-movement behavior.
 
 \begin{figure*}[!h]
  \centering
 \includegraphics[width=12cm, keepaspectratio]{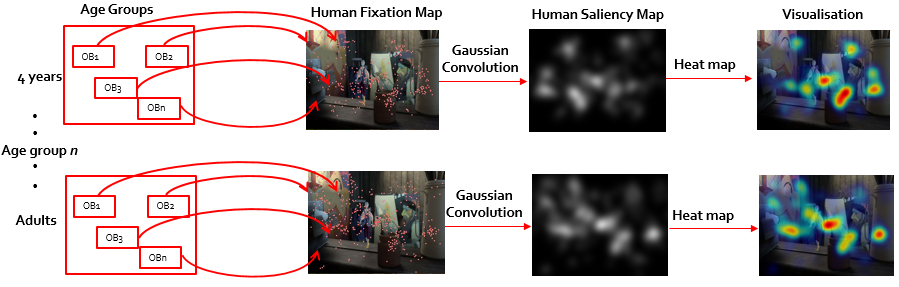}%
\caption{{\bf Map generation:}
Human fixation map and human saliency maps were generated by projecting all the fixations of all observers of an age group over an image. $OB_{n}$ stands for the $n^{th}$ observer of an age group.}
\label{fig1}
\end{figure*}

\section{Analysis}
In this section, we elaborate on our analysis to quantify the age-related differences in scene-viewing of observers. We develop measures to quantify three aspects of viewing behavior: explorativeness, agreement score within or across age groups and center bias, each of these contributes to the detailed understanding of how vision changes for scene viewing with age.\\

\subsection*{Explorativeness}

To evaluate eye movement behavior during scene exploration across age groups, we conducted an explorativeness analysis. Explorativeness was used to quantify the age-related differences in the distribution of gaze locations. As shown in Fig. 3, when participants in different age groups were observing the same set of images of our dataset, the set of least explored scenes were found to be different among observers belonging to different age groups. Thus, explorative behavior depends on the observer`s age as well.

\begin{figure*}[htp]
  \centering
 \hspace*{\fill}%
  \subcaptionbox{\label{fig2:a}}{\includegraphics[width=1.2in]{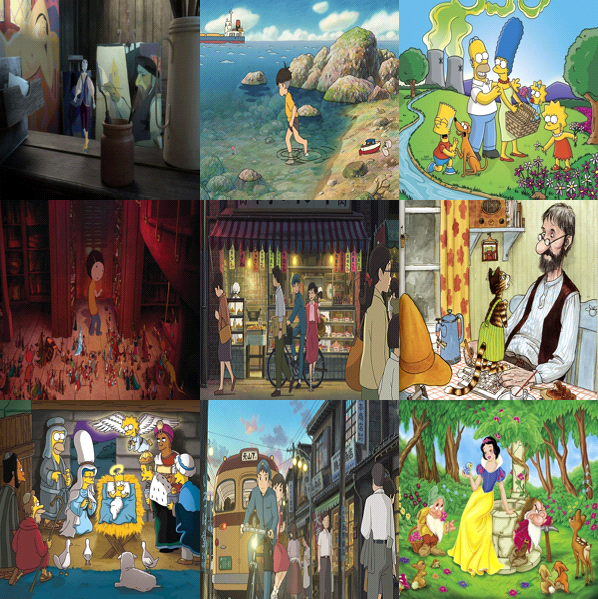}}%
  \subcaptionbox{\label{fig2:b}}{\includegraphics[width=1.2in]{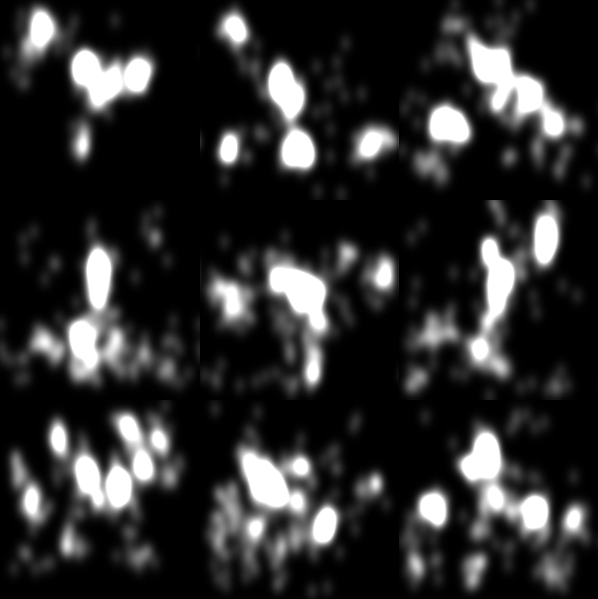}}%
 \subcaptionbox{\label{fig2:b}}{\includegraphics[width=1.2in]{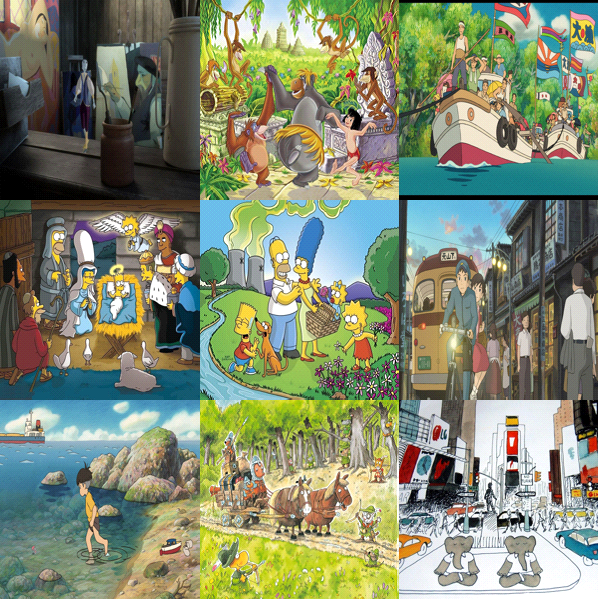}}%
\subcaptionbox{\label{fig2:b}}{\includegraphics[width=1.2in]{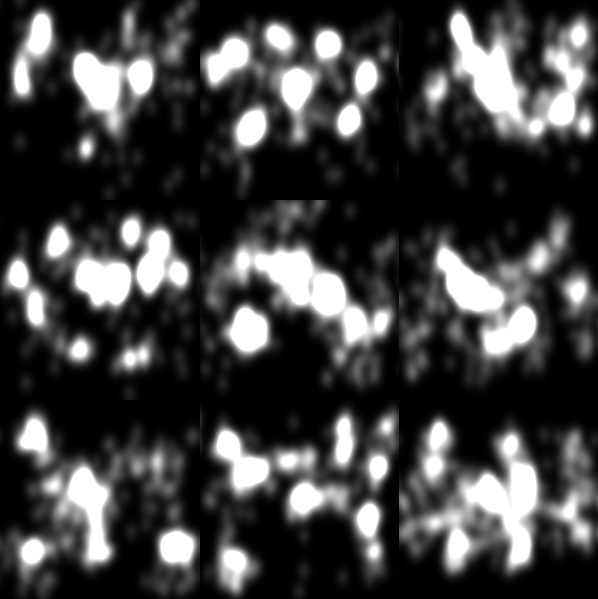}}%
\hspace*{\fill}%
\vfill
 
\caption{\textbf{Least explored images and their saliency maps :}(a,b) 4 year age group (c,d) young adult observers, the top 12 least explored images are ordered from the top left to bottom right}
 \end{figure*} 
 
 For any scene, we observed that a human saliency map differs between age groups. Thus, we analyzed explorative behavior of an observer across age groups. We calculated first-order entropy of the human saliency map to quantify the explorativeness of observers in a group. For the $i^{th}$ image of group $g$ it is computed as,
 
\begin{equation}
H(U^{g}_{i})=\displaystyle\sum_{l} h_{U^{g}_{i}}(l) * log (L\mathbin{/}h_{U^{g}_{i}}(l))
\end{equation}  
where $U^{g}_{i}$ is the human saliency map of the $i^{th}$ image from all observers in a group $g$ for which entropy is calculated and $h_{U^{g}_{i}}(l)$ is the histogram entry of intensity value $l$ in image $U^{g}_{i}$, and $L$ is the total number of pixels in $U^{g}_{i}$. 

    In the context of viewing behavior, a higher entropy corresponds to a more explorative viewing behavior by the observer, as their saliency points are more scattered in the given scene. Similarly, a lower entropy corresponds to less explorative behavior. The average behavior of each age group over all images was analyzed based on the average entropy.
   
   The results of the analysis suggested that: 
\begin{enumerate}
\item Explorativeness increases monotonically with age, $r(29)=0.99$, $p<0.001$. This can be seen in Fig. 4, which plots the entropy of all images for each age group.  The histograms of entropy of all images for different age groups are illustrated in Fig. 5.

\item  Adults had higher exploration tendency, which implies that during scene exploration, they tended to direct their gazes at different level of details in a given scene. On the contrary, being less explorative, children tended to direct their gazes towards fewer details of the scene. This is implied from the study in \cite{klkl}, which reported that decrease in image resolution i.e changing the level of detail is responded by the observers by decreasing the spread of the fixation landing i.e. entropy on the image.   

\item One-way ANOVA analysis showed that explorativeness varied significantly among the age groups, $F(3,29)=15.8$, $p<0.001$. Post-hoc analysis indicated that the changes in explorativeness score were significantly different between four to eight years, four to young adults, six to eight years, and six to young adult, all $p<0.01$. However, no difference was found between eight-years and young adults, suggesting that from the age of eight years explorativeness behavior is adult-like.

\end{enumerate}    

\begin{figure}[!ht]
\centering
\includegraphics[width=6cm,height=6cm,keepaspectratio]{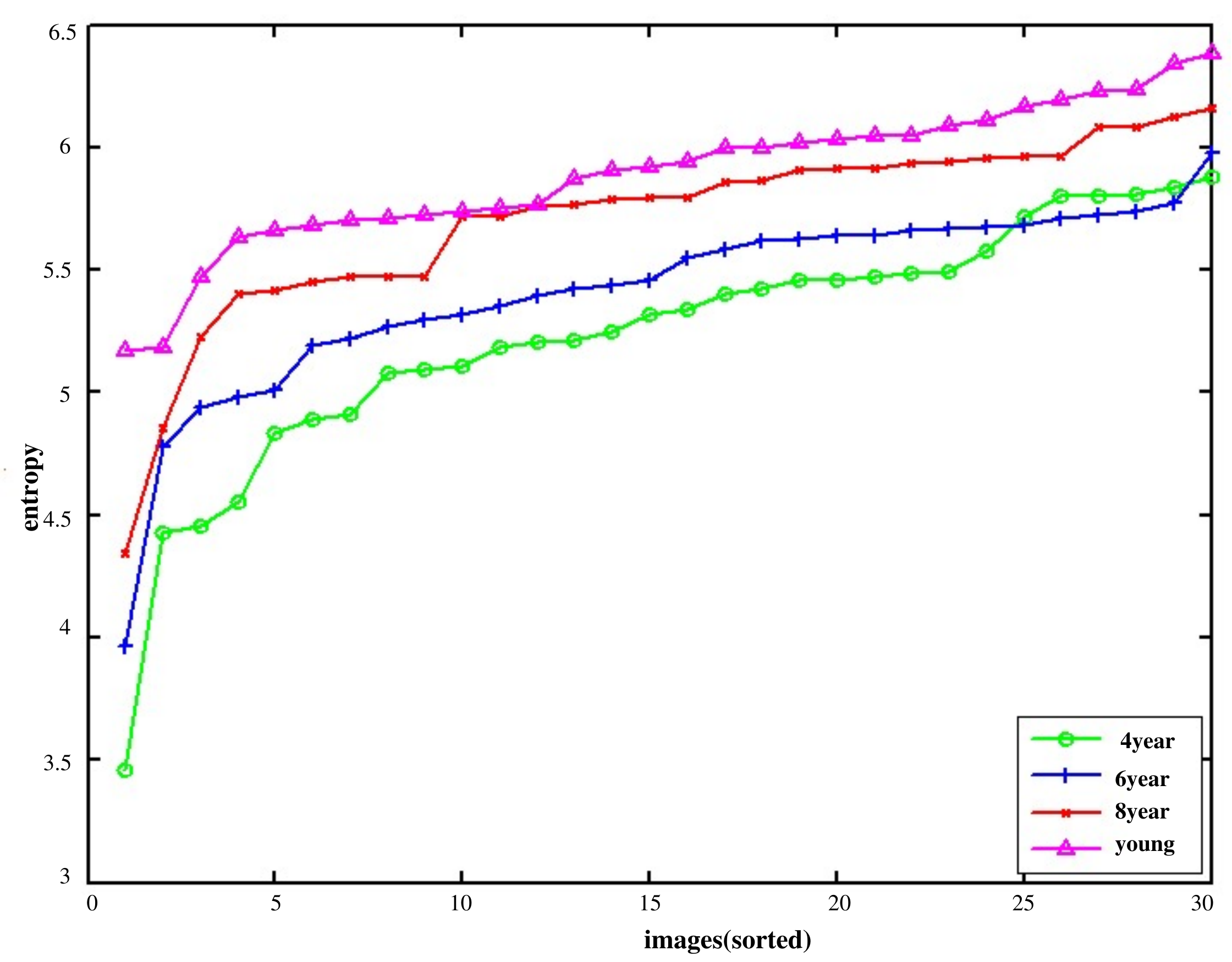}
\caption{Entropy plotted in sorted order for different age groups over all the stimuli}
\end{figure}

\begin{figure}[!ht]
\centering
\includegraphics[width=6cm,height=6cm,keepaspectratio]{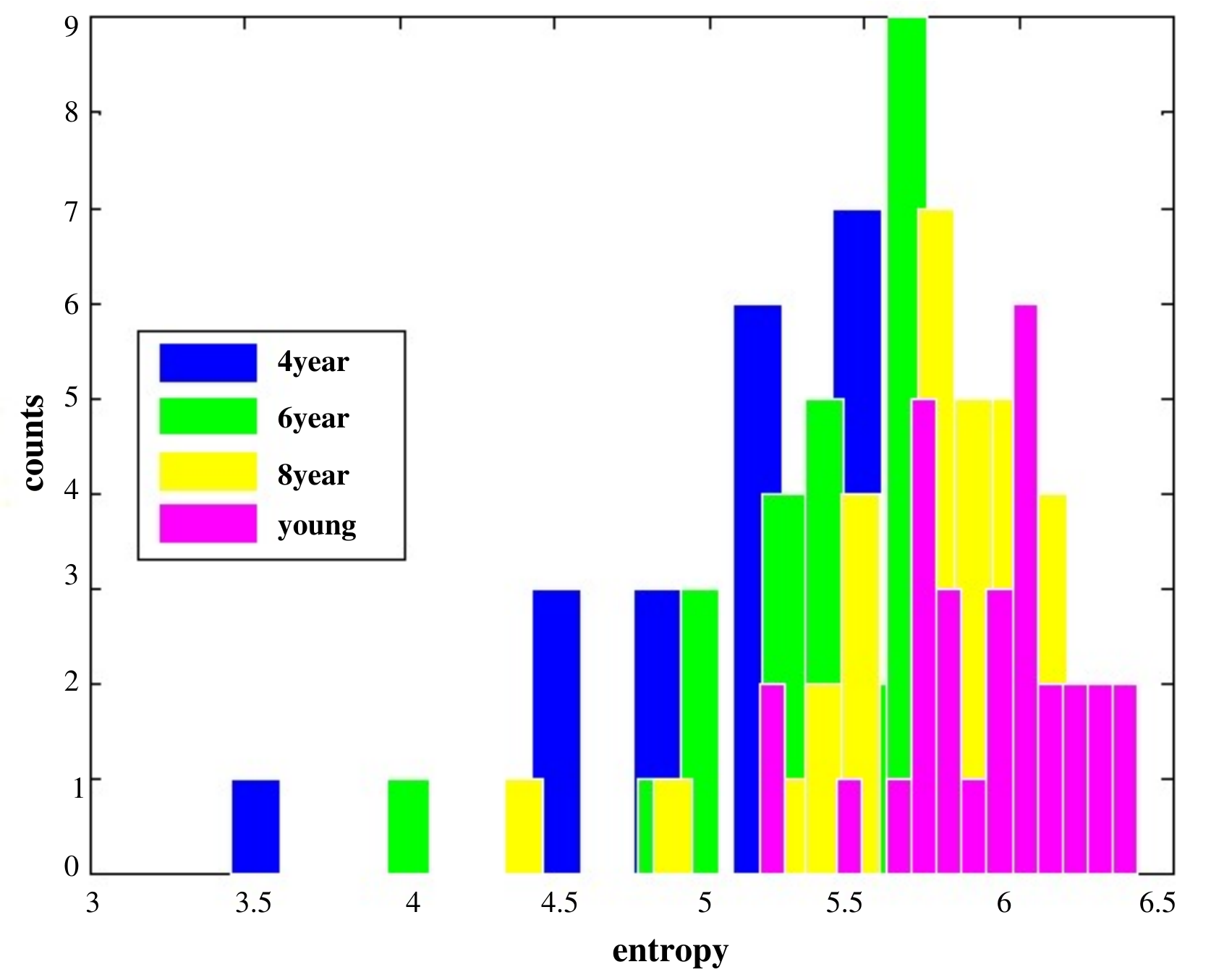}
\caption{\textbf{Histogram of entropy:-}The histogram of entropy indicates that there is a shift from left to right for 4 year to young adult age group}
\end{figure}

\subsection{Agreement analysis}

Explorativeness reflects the difference in the scene-viewing behavior of observers from different age groups. However, explorativeness score is unable to answer questions such as: do observers belonging to the same age group explore the same spatial regions of the image? And is there any agreement among observers in terms of explored regions across age groups? Explorativeness falls short of checking for similarity of explored regions within age groups and between age groups. It should be noted that poor agreement of fixation landings between adults and children leads to imprecise prediction when using  saliency models that are originally developed for adults. This motivated us to conduct an agreement analysis. 
    	 
   The area under the curve (AUC) is the most commonly used metric in the literature for discrete ground truth saliency maps \cite{p}, and we choose it for our analysis. The AUC-based measure analyzed how well the human saliency map of fixation points of all observers of an age group could be used to find the pooled fixation locations of all observers from the group, as well as observers from different groups. The age group of which the saliency map was used became the source group, and the group for which the fixation locations were being used as target group. Thus, under the intra-age group agreement analysis, the source and target belonged to the same group, and for inter-age group analysis, the source age group was different from the target group.

\begin{equation}
    TPR_{U_{n}}^{g_sg_t}(I_i)=\dfrac{TP_{U_{n}}^{g_sg_t}(I_i)}{TP_{U_{n}}^{g_sg_t}(I_i)+FN_{U_{n}}^{g_sg_t}(I_i)}
  \end{equation}
  \begin{equation}
    FPR_{U_{n}}^{g_sg_t}(I_i)=\dfrac{FP_{U_{n}}^{g_sg_t}(I_i)}{TP_{U_{n}}^{g_sg_t}(I_i)+FN_{U_{n}}^{g_sg_t}(I_i)}
  \end{equation}
  
Where the TPR for the $i^{th}$ image is the extent to which the fixation points of observers in group $g_t$ agree to the $n^{th}$ thresholded saliency map $U_{n}$ of observers from source group $g_s$. Similarly, FPR deals with non-fixation points that have been considered fixation points. The TPR and FPR for all $T$-thresholded saliency maps of an image were combined into a vector of $T$ dimension. The area under the ROC curve plotted between TPR and FPR gave us the AUC-score, and an average of these scores across all stimuli of the dataset provided the agreement score of the group.  
     
    The intra-age and inter-age group agreement accuracies were then calculated in terms of AUC-score. For intra-group analysis, $g_s$ and $g_t$ were the same, whereas for inter-group analysis, they were different. For a given image, this tells us how accurately the fixation locations of all observers in the group were covered under the differently thresholded saliency maps of observers from the same or different age groups. We can visualize the intra-age and inter-age group agreement results of analysis in Fig. 6.
   
\begin{figure*}[!ht]
  \centering
 
 \subcaptionbox{\label{fig3:a} \textbf{Intra-age group:} target fixation points of 4 year by source saleincy map of 4 year.}{\includegraphics[width=6cm,height=4cm,keepaspectratio]{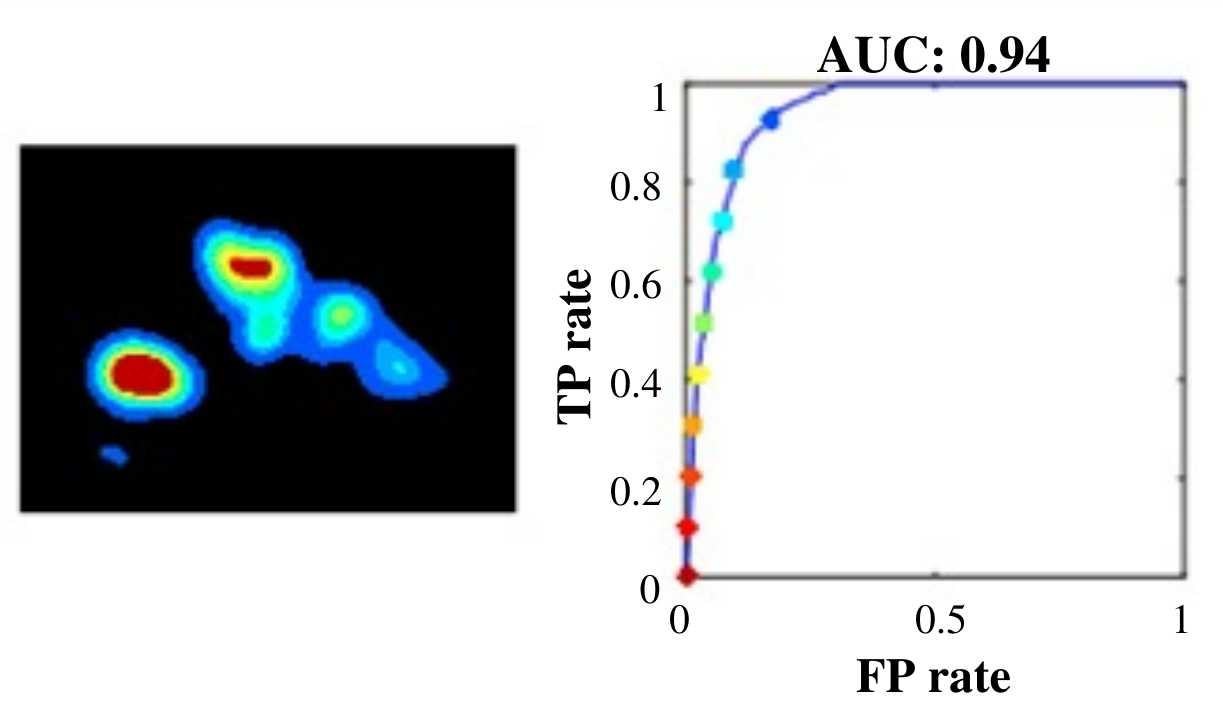}}\hspace{1em}%
 \subcaptionbox{\label{fig3:a} \textbf{Intra-age group:} target fixation points of young by source saleincy map of young.}{\includegraphics[width=6cm,height=4cm,keepaspectratio]{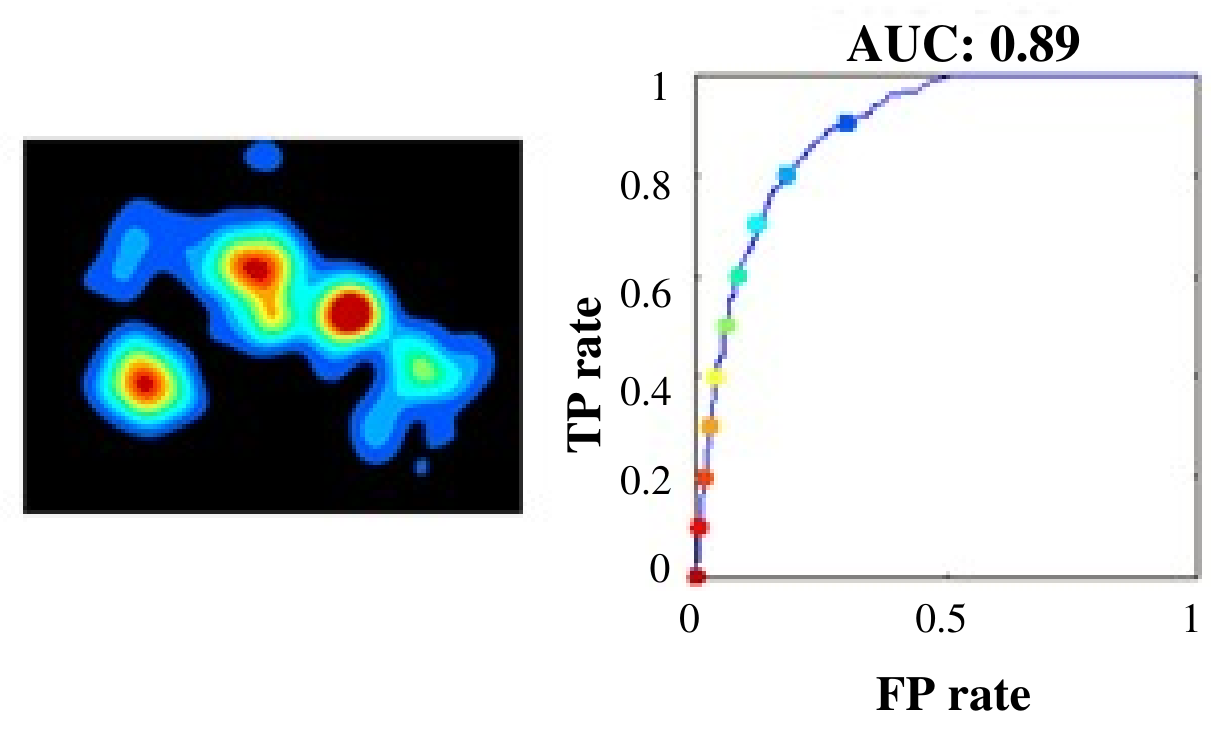}}\hspace{1em}
  \bigskip
 
 \subcaptionbox{\label{fig3:a} \textbf{Inter-age group:} target fixation points of 4 year by source saleincy map of young.}{\includegraphics[width=6cm,height=4cm,keepaspectratio]{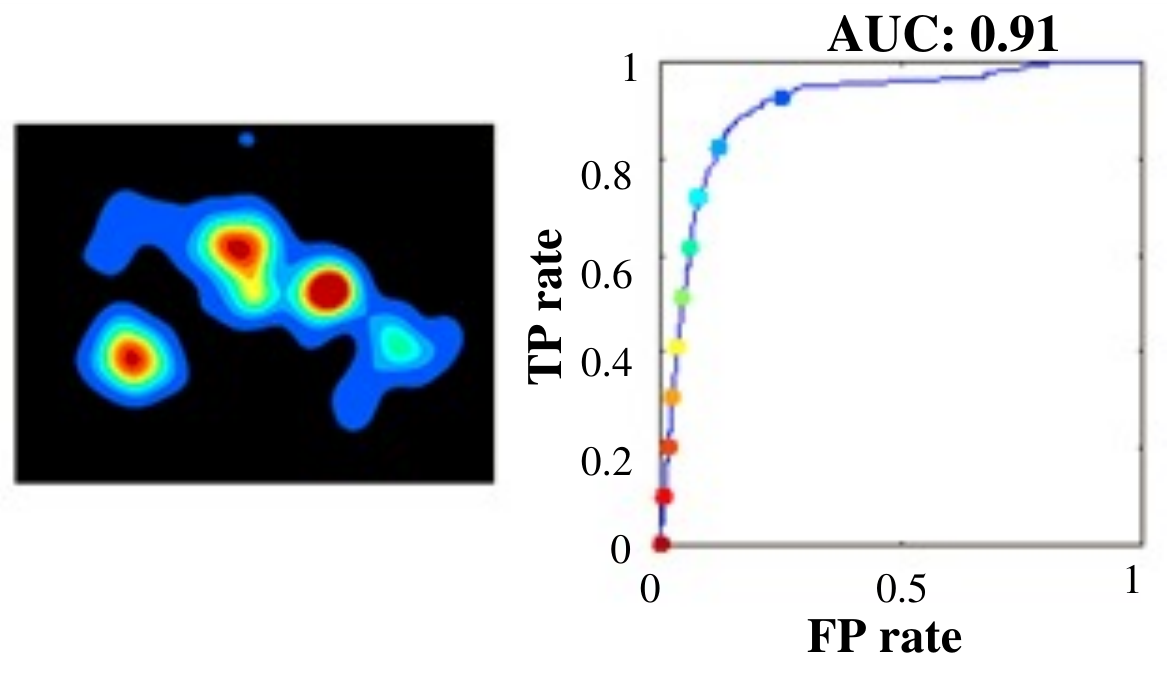}}\hspace{1em}
 \subcaptionbox{\label{fig3:a} \textbf{Inter-age group} target fixation points of young by source saleincy map of 4 year.}{\includegraphics[width=6cm,height=4cm,keepaspectratio]{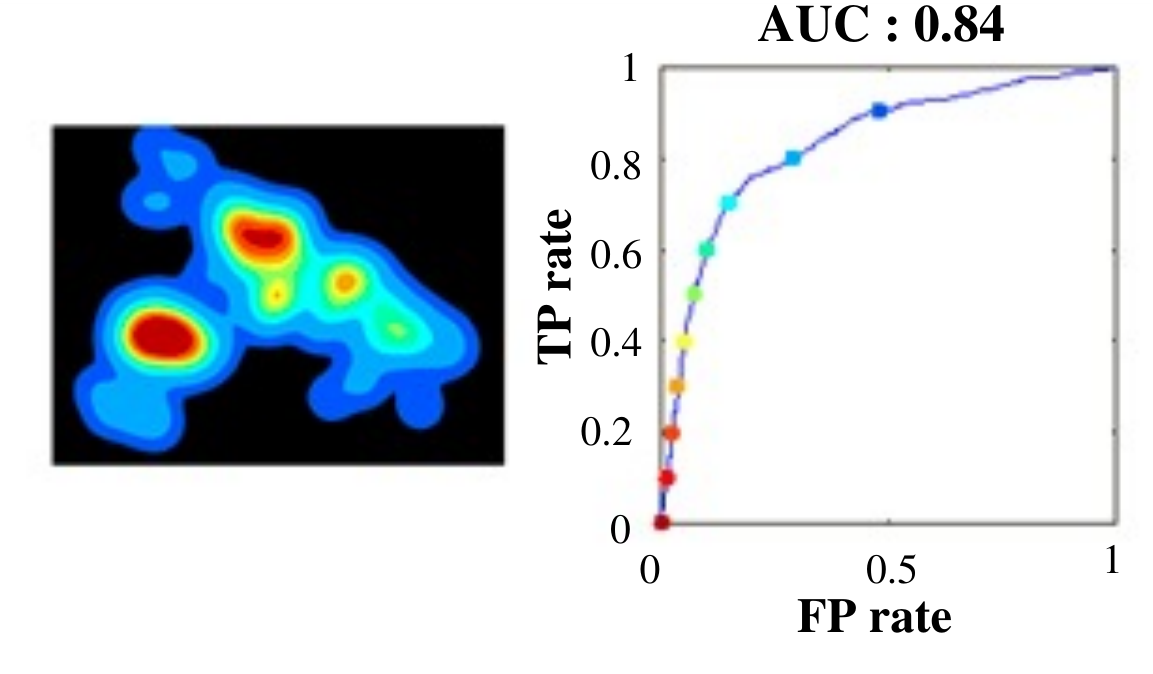}}\hspace{1em}

\caption{\textbf{Agreement score analysis:} The heat map visualizes agreement behavior in predicting the target fixation points by source saliency map and the ROC calculates the quantitative value of the agreement score.}
 
\end{figure*}


   The key suggestions from the intra-age and inter-age group agreement analysis are as follows:\\
    \textbf{Intra-age group agreement analysis:}
    \begin{enumerate}
  
    \item  As shown in Fig. 7, the average agreement score of the four-year age group was highest for all images across age groups. The score started decreasing as observer's age increased up to 8 year age group by showing a strong negative correlation, $r(29)=-0.88$, $p<0.001$, but, similar to the explorativeness results, the agreement score suggested that scene-viewing tendency matures at the age of eight.
\item  Comparing the results of explorativeness and agreement analysis, it is interesting to note that the trend followed by the intra-group agreement analysis was opposite to exhibited by the explorativeness results. This makes sense: as explorativeness decreased, observers tended to focus on lesser details of the scene, mostly the ones that were the key areas of the image. This suggests that the fixation points of the observers of least explorative age group would mostly be consistent with one another, and would be mostly localized at key objects and, hence, the agreement score would be high.
\item One-way ANOVA test suggested the age impacted on explorativeness tendency   			 $F(3,29)=65.8$, $p<0.01$. As shown in Fig. 8, 8 years and adults have significantly less intra-age group agreement than 4 and 6 year olds, $p<0.01$. This can be understood by the fact that 8 year olds and adults are the most explorative, and there salient regions may not be consistent with one another at higher level of the details.    
\end{enumerate}
 
\begin{figure}[!ht]
\centering
\includegraphics[width=6cm,height=6cm,keepaspectratio]{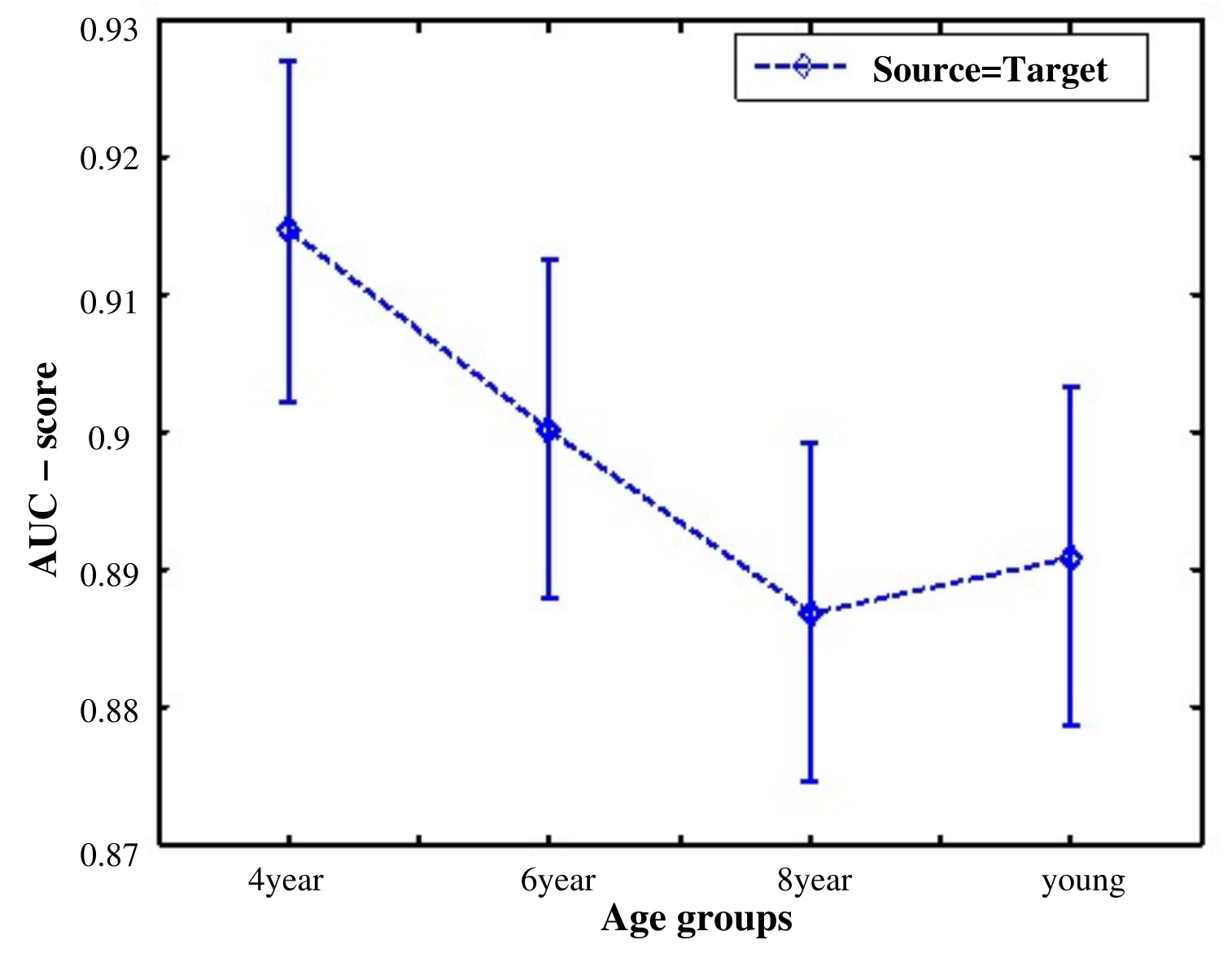}
\caption{Intra-age group agreement scores, which reflects that kids agree more in explored locations than younger adults }
\end{figure}       
      
\begin{table*}[]
\large
\centering
\caption{\textbf{Agreement score}: Average agreement score of human saliency map of observers from the source group in predicting fixation points of target age group.}
\begin{adjustbox}{max width=\textwidth}
\begin{tabular}{|l|c|c|c|c|}\hline
\backslashbox{Source}{Target}
&\makebox[2em]{4 years}&\makebox[2em]{6 years}&\makebox[2em]{8 years}
&\makebox[2em]{adult}\\\hline\hline
4 year&\cellcolor{green}0.9148&0.8756&0.8695&0.8683\\\hline
6 year&0.8463&\cellcolor{green}0.9003&0.8509&0.8493\\\hline
8 year&0.8150&0.8269&\cellcolor{green}0.8870&0.8343\\\hline
adult&0.8122&0.8265&0.8340&\cellcolor{green}0.8910\\\hline
\end{tabular}
\end{adjustbox}
\end{table*}

\textbf{Inter-age group agreement analysis:}
    \begin{enumerate}
    \item Table 2 shows that the agreement scores of inter-age group experiments was lower than that of intra-age group experiments for all ages. Thus, it was even more evident that age has an impact on visual behavior as the same age group maps defined the fixations more precisely.
\item The most important contribution of the inter analysis was that the saliency map of adult subjects showed the poorest performance in predicting the fixation points of the other age groups as shown in the fifth column of Table 2: Agreement score of adults predicting all others is significantly less than the agreement scores of diagonal colored boxes of the table (prediction by same age groups). One-way ANOVA analysis indicated significant differences in performance of adults predicting 4 year, 6 year, and 8 years than the prediction by the same age-group, $F(3, 29)=7.49$, $p<0.03$. Thus, ignoring the age factor and using conventional models developed and learned over young adults can not give optimal performance for other age groups. This calls for the modification of existing models to make them adapt to age. Fig 8 shows the comparison of agreement score for saliency maps of four year olds and young adults in finding the target fixations of different age groups.
\end{enumerate}

\begin{figure}[!ht]
\centering
\includegraphics[width=6cm,height=6cm,keepaspectratio]{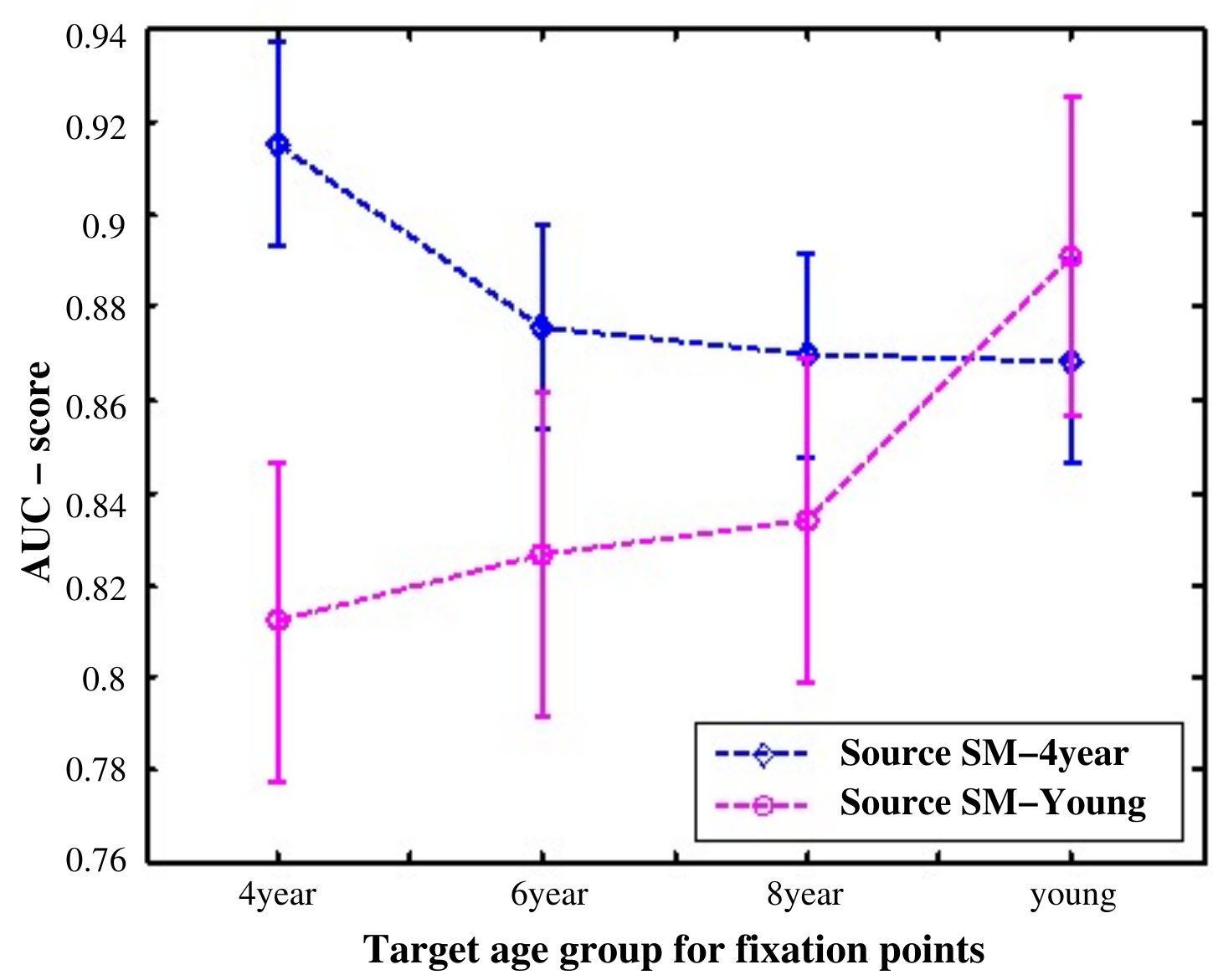}
\caption{Agreement analysis results for the source saliency map of four years and adults in finding the target fixation of different age groups}
\end{figure}

\subsection{Center bias}
The term ``center bias" has been studied using the eye-tracking techniques, and it reflects the human tendency of looking at the center of a given image \cite{q}. A possible explanation of this tendency lies in the fact that while taking pictures, photographers tend to keep the region of interest at the center of the frame, i.e., photographer bias. Due to the photographer's bias, human observers develop the tendency of focusing on the center of a given scene to obtain maximum information while exploring: this is called the observer's bias. Studies have established the existence of the center bias, but only a few scholars have considered the center bias in their computational models \cite{n}\cite{mtmt}.

 The center bias greatly influences our viewing behavior but, to the best of our knowledge no study has investigating the age related differences in tendencies toward center bias in different groups.  The focus of our study is to reveal differences in center bias across age groups. We first calculated the average saliency map across all images for each age group, i.e., the center map. We then used this center map to measure agreement scores with fixation locations for all images across age groups. 
 
  As shown in Fig 9, age-related differences in center bias tendency suggested that the four year age group had the highest center bias among all age groups. It decreased with increasing age, where adult-like observation behavior was exhibited at 8 years of the age. The results of One way ANOVA analysis indicated that the any two age groups were significantly different, $F(3, 29)=8.15$, $p<0.03$. Further post hoc analysis indicated that both adults and 8 year are significantly different from 4 year and 6 year age groups,$p<0.01$.

\begin{figure}[!ht]
\centering
\includegraphics[width=7cm,height=7cm,keepaspectratio]{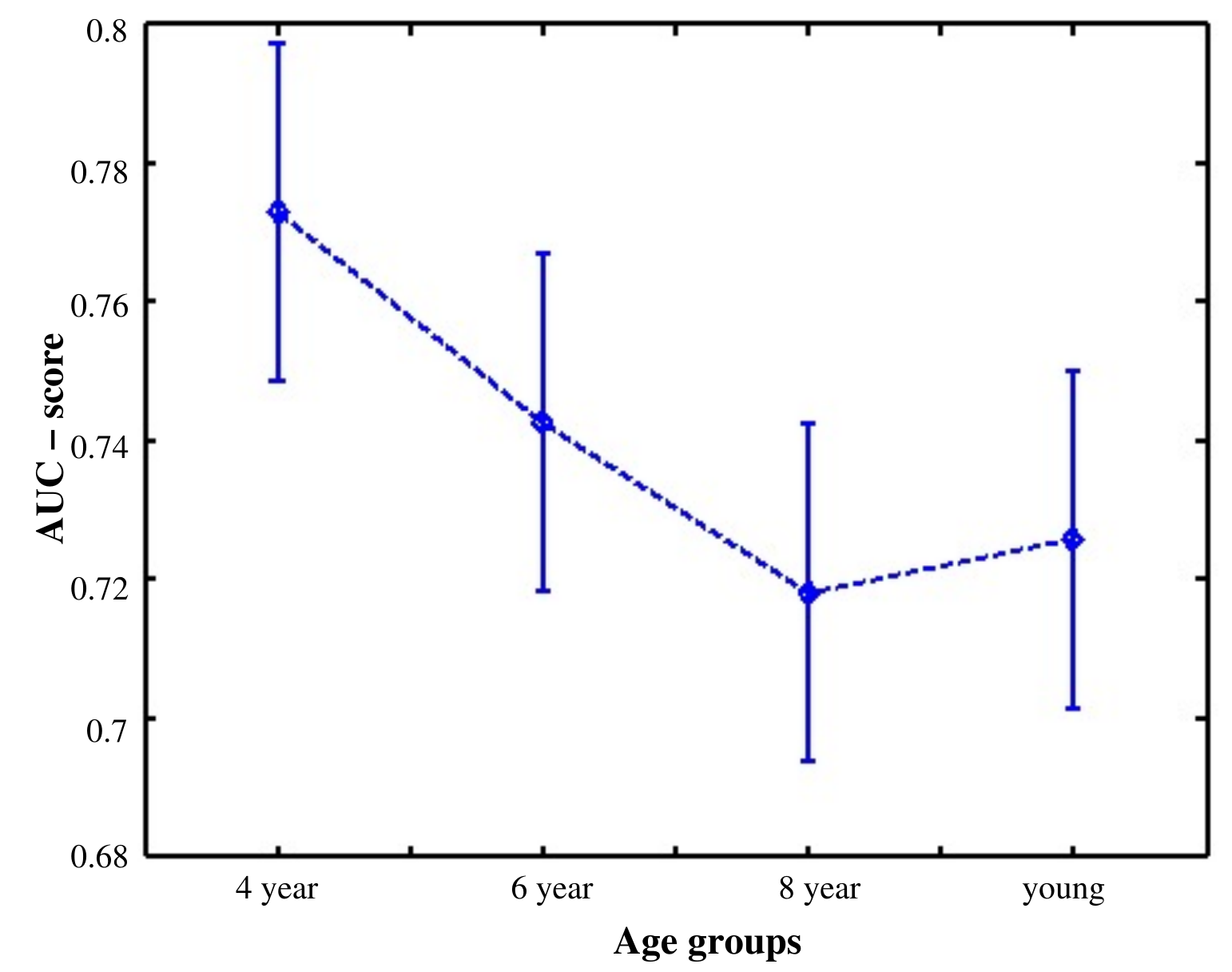}
\caption{Age-based changes in center bias tendency across age groups}
\end{figure}

\subsection{Results of analysis: A framework for the age-adapted model}
    
    We briefly present three main findings that helped us build the age-adapted computational model in the next section:
    \begin{enumerate}
    \item Results of Explorativeness anaysis suggested that children (four and six years) exhibited the least explorative behavior among the age groups. Explorativeness has a direct relation with the level of details an observer tends to explore \cite{klkl}. Age associated variation in explorativeness indicates that observers of different age groups viewed different levels of detail within a scene. This helped us to choose the scales of features extracted from the images to generate a master saliency map. The features scale selection should be such that they are capable of representing age-based variations at the level of detail of the observer. 
    \item Intra-group agreement scores were higher than inter-group agreement scores for various combinations of groups. This suggests that while training the model, it is advisable to train the model of a particular age group by using the fixation-map data of the same age group rather than the generalized fixation map data of young adults.
    \item The magnitude of the center bias was different among age groups. Thus, while including the center bias in age-adapted saliency model, we need to consider the age-related differences in center bias tendency.  
    \end{enumerate}

\section{The saliency model adapted to age}
Several computational models for visual saliency have been developed in past work to provide important insights into the underlying mechanisms of the human visual system. All existing models involve learning to predict regions of interest in images by considering the gaze behavior of young adults. Thus, these models are optimized to predict fixations of young adults, but at the same time, prediction accuracy of these models are not optimal for other age groups. Given the varying viewing behavior of observers belonging to different age groups as highlighted in the last section, provide us opportunity to optimize the prediction performance of existing modes for other age groups as well.

 We introduced a basic framework for age-adapted saliency models in the last section using the results of our analysis. This framework can be used to upgrade the state-of-the-art computational models to enable their predicted saliency maps to reflect the age related differences more accurately in scene-viewing behavior.

 In the proposed work, our age-adapted framework was tested with two types of computational models \cite{l}, \cite{fabab}. We chose these models carefully in light of the fact that they had different modeling architectures. We verified that the proposed age-adapted framework was generalizable, and could be applied to any type of existing model as, most of them follow the same basic structure with minor variations. The models chosen were the following: 

\begin{enumerate}

 \item The Itti's model \cite{l}, where different visual features are extracted over multiple scales of the input image and a saliency map is obtained by linearly integrating these feature maps into one. The proposed age-adapted framework was incorporated with this model by applying the multi-scale feature subset selection mechanism with a different set of optimal weights of feature integration learned over our age specific gaze dataset.
 \item A patch-based, age-adapted model was inspired by the patch-based method of saliency predictions \cite{fabab}, where the aim is to detect the saliency of the scene based on dissimilarity among neighbouring patches. The existing model was modified and the age-adapted framework was applied by varying patch size and the age-adapted weighting factor for the center bias. 

\end{enumerate}

The selection of these models relies on the fact that most of the bottom-up computational models follow this basic structure.

  \subsection{Age-adapted multi-scale feature subset selection and optimization based model}
Most existing bottom-up models follow the basic multi-scale feature selection architecture proposed by Itti et al \cite{l}. In these models, we observe the following basic structure: (a) Basic visual features such as color, intensity, and orientation, are extracted over multiple scales of the image, where each scale represents a different level of detail in the scene. (b) All features are investigated in parallel, to obtain the conspicuity map for each feature channel. (c) These features are integrated to obtain the saliency map. 

There are three concerns in developing an age-adapted model over this basic structure of saliency prediction - First, we need to choose the appropriate set of feature scales for different age groups, as our results suggest that different age groups tend to explore different levels of detail in scenes. Second, we need to include the center bias in the proposed model by considering the fact that the strength of the center bias varies with observer age. Third, we need to combine the extracted features over an optimized set of weights for different age groups. This optimization is achieved through a supervised way of learning weights for different age groups. 

\textbf{(a) Multi-scale feature subset selection: Proposed S+C}\\
  We used the multi-scale feature extraction technique proposed in the famous Itti et al.'s saliency model \cite{l}. The different scales represented the different levels of detail in scenes, from finer details to coarser object-level details. As stated earlier for more explorative observers, all levels of details were important and, hence, all feature scales were used to learn the model; for the less explorative observers, only coarser-level details were important and, so, only a few scales sufficed. 

Observers from different age groups showed different levels of explorativenss. Thus, to make our model adapt to age-related differences in scene viewing behavior, we focused on a feature scale selection mechanism, where we identified the subsets of the feature maps that best represented the different levels of details viewed by the observers of different age groups.

\begin{table*}[]
\large
\centering
\caption{\textbf{Average prediction accuracy by multi-scale feature subset selection only:} where as a$\sim$b means from scale a to scale b are selected (Proposed S+C). Scale 6 is the coarsest level and scale 1 is the finest}
\begin{adjustbox}{max width=\textwidth}
\begin{tabular}{|l||*{13}{c|}}\hline
\backslashbox{Age}{Scale}
&\makebox[2em]{1$\sim$6}&\makebox[2em]{2$\sim$6}&\makebox[2em]{3$\sim$6}
&\makebox[2em]{4$\sim$6}&\makebox[2em]{5$\sim$6}&\makebox[2em]{6$\sim$6}\\\hline\hline
4 year & 0.7074 & 0.7180 & 0.7288 & 0.7280 & \textbf{0.7381} & 0.7366  \\\hline
6 year & 0.6839  & 0.6940 & 0.6978 &\textbf{0.7183}  & 0.7044  & 0.7071  \\\hline
8 year & 0.6640  & 0.6655 &  \textbf{0.6722}  & 0.6664  & 0.6572  & 0.6566  \\\hline
adult & \textbf{0.6628}  & 0.6573 & 0.6541 & 0.6512  & 0.6470 & 0.6495  \\\hline
\end{tabular}
\end{adjustbox}
\end{table*}

We now discuss the steps to extract features for our age-adapted saliency model. For an input image, eight spatial scales were first developed using a Gaussian pyramid. The features were then extracted using the ``center-surround" operations with the same settings as in \cite{l} to yield six intensity maps $\mathcal{I}_{i}$, 12 color maps - six for $\mathcal{RG}_{i}$ and six for $\mathcal{BY}_{i}$ each and 24 orientation maps - $\mathcal{O}_i(\theta)$ i.e., sets of six maps computed for four orientation $\theta \in \{0,45,90,135\}$. The 6 maps for different feature represents different level of detail in scene..  
The Feature maps were then combined into three ``conspicuous maps", $\bar{\mathcal{I}}$ for intensity, $\bar{\mathcal{C}}$ for color, and $\bar{\mathcal{O}}$ for orientation. However, as stated above, unlike Itti et al.'s model, this point-wise combination was not conducted over all six maps; we also chose subsets of six maps for each age group. The point wise combination of feature map was: 
 
\begin{equation}
  \bar{\mathcal{I}}=\bigoplus_{i=s}^6 \mathcal{N}(\mathcal{I}_i)
\end{equation}
\begin{equation}
  \bar{\mathcal{C}}=\bigoplus_{i=s}^6 [\mathcal{N}(\mathcal{RG}_i)+\mathcal{N}(\mathcal{BY}_i)]
\end{equation}
\begin{equation}
  \bar{\mathcal{O}}=\displaystyle\sum_{\theta \in \{0,45,90,135\}}\bigoplus_{i=s}^6 \mathcal{N}(\mathcal{O}_i(\theta))
\end{equation}
where $\mathcal{N}$ represents the normalization and $s$ is the starting index from where maps were taken.

We developed six cases by varying s to 1,2,3,4,5, and 6. If s = 1, the subset of feature scale starting from scale 1 (finer) to scale 6 (coarser) had to be combined. Similarly if s = 6, only the feature scale 6 was used. Without using the trend toward explorativeness found in the analysis section, we evaluated the model over all such possible subsets for all groups, and defined the subset for each age that best represented the gaze levels (finer to coarser) of the observers in a given age group. 

 As shown in prediction results in Table 3, younger age groups (4 years, 6 years, and 8 years)  are performing better than adults, which makes use of all scales (1$\sim$6) however, the prediction accuracy of children were not optimized on the existing scale ($1\sim6$). The predictive performance of children get optimized if used coarser scales and ignore finer ones (as in Table 3, scele $5\sim6$, $4\sim6$, and $3\sim6$ are optmized scale selection for 4 year, 6 year, and 8 year age groups respectively) while for young adults, prediction accuracy was higher if we chose all scales (similar to \cite{l}). It is interesting to note that this result is consistent with our earlier results, i.e., children are less explorative than young adults and, hence, require only coarser scales to predict their fixations. Age related differences in center bias tendency was also incorporated in this model by including a differently weighted center-map as explained in     following section.

\textbf{(b) Training and Testing: Feature Combination Optimization: Proposed S+I+C} \\
In this section we proposed another modifcation in existing models based on our second recommendation reported in Results of analysis: a framework for the age-adapted model section. The choice of linear integration of feature maps used in previous section was ill-suited because different features contribute differently to the final saliency map. 

  Some state-of-the-art models address \cite{n} this by learning the optimal weights of feature integration in a supervised manner. These optimal weights are, however, not suitable for our age-adapted mechanism, as they are learned only over eye-tracking data collected for young adults. To fit this into our scenario, we learn these optimal weights over features extracted from age-specific subsets of the dataset. 
We divided the dataset into a training set with 20 images and a test set with the remaining images. Color, intensity, and orientation features were extracted for the training images. We then selected $P$ strongly positive and negative samples, each corresponding to the top and least-rated salient locations of the human saliency map of all observers generated from ground truth eye-tracking  data. 
  
  Our analysis of the agreement scores of the prediction of the fixation point suggests that intra-age group fixation point prediction was better than inter-age group performance. In other words, the fixation points of the observers were better predicted by the saliency maps of observers of the same group rather than those of observers of other groups. Thus, the $P$ positive and negative samples to be chosen were age group specific, i.e., the positive and negative samples for all age groups were differently chosen for training. 
 
  We fixed value $P$ to 10; choosing more samples only involved adding redundancy and yielded no performance improvement. For a given set of features and labels (positive and negative samples) for an age group, liblinear SVM was used to learn the model parameters to predict salient locations on the training images. Thus, we obtained model parameters for predefined features over all age groups.
For a given test image, we first collected its features as described in the multi-scale feature selection mechanism, and further predicted saliency values at each pixels as,
\begin{equation}
  S_g(I_i)=w_gX^T(I_i)+b_g
\end{equation}
  where $w_g$ and $b_g$ are model parameters learned for each age group $g$ and $X(I_i)$ is feature vector for the $i^{th}$ test image, this vector is composed of intensity ($\bar{\mathcal{I}}$), color ($\bar{\mathcal{C}}$), and orientation ($\bar{\mathcal{O}}$) features. Based on the saliency values we classified the local pixel as salient or not.
  
  Integrating the feature maps over the optimally set weights learned over the age-specific dataset suggests further improvement in prediction accuracy for all age groups including young-adults, as shown in Table 4. Age-related differences in center bias tendency were also considered while evaluating the performance of the proposed model. The method of incorporating center bias in the age-adapted model is explained in following section. The improvement in prediction performance for our proposed S+I+C model is shown in Fig. 10.
           
  \begin{table*}[]
\large
\centering
\caption{Average prediction accuracy by combining scale based subset selection, nonlinear integration and age-adapted center bias (Proposed S+I+C).}
\begin{adjustbox}{max width=\textwidth}
\begin{tabular}{|l||*{13}{c|}}\hline
\backslashbox{Age}{Scale}
&\makebox[2em]{1$\sim$6}&\makebox[2em]{2$\sim$6}&\makebox[2em]{3$\sim$6}
&\makebox[2em]{4$\sim$6}&\makebox[2em]{5$\sim$6}&\makebox[2em]{6}\\\hline\hline
4 year & 0.7387  & 0.7409 & 0.7374 & 0.7414 & \textbf{0.7434} & 0.7388 \\\hline
6 year & 0.7203 & 0.7218 & 0.7155& 0.7201  &\textbf{0.7295}  & 0.7160  \\\hline
8 year & 0.6766  & \textbf{0.6833} & 0.6698 & 0.6590  & 0.6655 & 0.6728  \\\hline
adult & \textbf{0.6646}  & 0.6637 & 0.6613 & 0.6549 & 0.6533  & 0.6585 \\\hline
\end{tabular}
\end{adjustbox}
\end{table*}

  \textbf{(c) Age-adapted model for center bias} \\
  Humans have the tendency to observe at the center of a given scene. This behavior can be incorporated with existing saliency models by simply defining saliency to include weight factor $C$, which is inversely propositional to the distance to the center of the pixel under consideration.  

\begin{equation}
  C(i)=1- d(c, p_i)/D
\end{equation}
where $d(c, p_i)$ is the distance between the pixel under consideration $p_i$ and center pixel $c$ and $D$ is the maximum distance used as a normalization factor. Further center bias $C(i)$ is updated based on the results of analysis reflect the age-related variations. $w_kC(i)$ is the updated center bias weight factor, where $w_k$ is the strength of the center bias tendency for different age groups. 

   \subsection{Age-adapted patch based saliency model: Proposed P}
      Another approach that we choose to verify the generalizability of our age-adapted framework is the patch-based model for saleincy prediction \cite{fabab}. This technique follows the given basic structure: (a) Image is first divided into patches of the same size. (b) The set of features are extracted from these patches. (c) Finally, the spatial dissimilarity among neighbouring patches is evaluated to generate the saliency map. 
   
As pointed out earlier, we do not use this model as is, but introduce some modifications. For this, we represent different features extracted from a patch by using the subset of eigenvalues obtained after SVD decomposition of the feature matrix. We elaborate this before explaining how to render this newly constructed model age-adapted.
    
       \textbf{(i) SVD decomposition based representation of features} \\
    We first construct the feature matrix.  The first step in feature matrix construction is to extract non-overlapping patches of size $t\times t$ from a given image I of size $M \times N$. Thus, the total number of patches $n_{p} = M \times N \slash t \times t $. Further, each patch is represented by a column vector of features $f_i$, where $i$ indexes the patch. $f_i$ is obtained by combining three color of features ($L^{*}, a^{*}, b^{*}$) and two intesity features ($I_x, I_y$). This generates a feature vector for each patch that appears as $[L_1,L_2,....,L_t,a_1,a_2,....,a_t$, \\ $b_1,b_2,....,b_t,
  I_{x_1},I_{x_2},..I_{x_t},I_{y_1},I_{y_2},...,I_{y_t}]$. 
    Finally, feature matrix $X$, $X=[ f_1, f_2, ....,f_{n_{p}}]$ for the entire image is obtained by combining the feature vectors of all patches \\

 Once the feature matrix representation is ready, we generate the covariance matrix representation of feature matrix $X$,  $C = X'X^T$.  Principle component analysis was used to diagonalizes covariance matrix $C$ by solving the following eigen vector problem:
\begin{equation}
  \lambda V = CV
\end{equation}
where $V$ are the eigen vectors of C and $\lambda$ represents the corresponding eigenvalues. The eigenvectors are ranked in descending order of eigen values. Choosing $d$ eigenvectors corresponding to the $d$ largest eigenvalues gives us the basis along the directions of maximum variance in features. Thus, the resultant matrix can be represented as $E=[V_1, V_2, ..V_d]^T$.

\textbf{(ii) Saliency measurement} In the final step, saliency can be measured based on the dismilarity between patches, which can be simply defined as Euclidean distance between patches in reduced dimension. 
 \begin{equation}
  S(R_i)=\omega(i)\sum_{j=1}^{L} \frac{\sum_{s=1}^{d}|x_{si}-x_{sj}|}{1+dist(p_i, p_j)}
\end{equation}
where $i,j$ are the $i^{th}$ and $j^{th}$ patches of an image and $\omega(i)$ can be defined as a weight factor to adjust the center bias.     

   \begin{table*}[]
\large
\centering
\caption{Average prediction accuracy by proposed patch based method, for different scale 1,2,3,4 (Proposed P).}
\begin{adjustbox}{max width=\textwidth}
\begin{tabular}{|l||*{9}{c|}}\hline
\backslashbox{Age}{Scale}
&\makebox[2em]{1$\sim$4}&\makebox[2em]{2$\sim$4}&\makebox[2em]{3$\sim$4}
&\makebox[2em]{4$\sim$4}\\\hline\hline
4 year & 0.7678  & 0.7767 & \textbf{0.7773} & 0.7772 \\\hline
6 year & 0.7400 & \textbf{0.7483} & 0.7480 & 0.7482 \\\hline
8 year & 0.7195  & \textbf{0.7279}  & 0.7272& 0.7269 \\\hline
adult & \textbf{0.7212} & 0.7113 & 0.7208 & 0.7188 \\\hline
\end{tabular}
\end{adjustbox}
\end{table*} 
  
    Similarly to the previous model, the age-adapted framework is incorporated into this model by selecting a different subset of patch sizes for different age groups and incorporating the age-adapted center bias.  We can select patch sizes from the set $\left \{8, 16, 32, 64\right \}$, which varies from finer to coarser scale. The result of this model is shown in Table 5. As expected, all scales are suitable for young adults, whereas children are more sensitive to fewer scales.
    
    Table 6 lists the fixation prediction accuracies of some famous existing saliency models executed unaltered for our age specific gaze dataset over observers of different age groups. From Fig. 10 and Table 6, it is clear that our modification of Itti et al.'s model and patch-based models that leverage the age-adapted framework outperformed existing models. Our modified in patch based improves the prediction performance for adult observers as well. We believe that difference in the fixation prediction accuracies was evidence for the fact that our algorithm not only personalizes saliency models to achieve optimal performance according to the observer's age group but also improves the prediction performance of adults.

\begin{figure}[!ht]
  \centering
 
 \includegraphics[width=9cm,height=9cm,keepaspectratio]{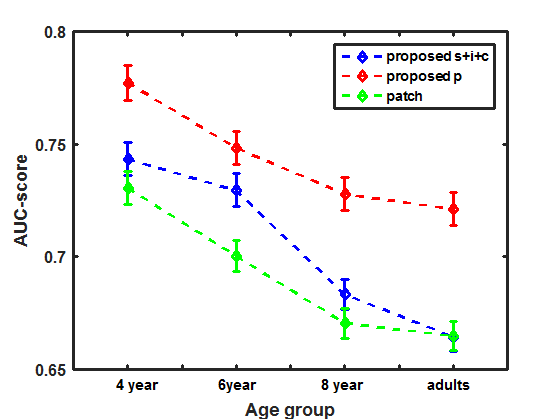}
 \caption{Comparison of age-adapted proposed saliency models with baseline models of computational attention system}
\end{figure}

   \begin{table*}[]
\large
\centering
\caption{Comparison table of our proposed (S+I+C and P models) age-adapted models with available computational models of saliency prediction.}
\begin{adjustbox}{max width=\textwidth}
\begin{tabular}{|l||*{6}{c|}}\hline
\backslashbox{Age}{Model}
&\makebox[4.1em]{S+I+C}&\makebox[4.1em]{P}&\makebox[4.1em]{\text{Itti's \cite{l}}}
&\makebox[4.1em]{\text{GBVS \cite{a4}}}&\makebox[4.1em]{\text{Judd's \cite{n}}}&\makebox[4em]{\text{Patch \cite{fababa}}}\\\hline\hline
4 year & 0.7434  & \textbf{0.7773} & 0.6218 & 0.7184& 0.7296& 0.7306 \\\hline
6 year & 0.7295 & \textbf{0.7483} & 0.6147 & 0.6969& 0.7033& 0.7003 \\\hline
8 year & 0.6833  & \textbf{0.7279}  & 0.6027 &0.6722 &0.6721&0.6706\\\hline
adult & 0.6646  & \textbf{0.7212} & 0.6062 & 0.6707&0.6660&0.6649 \\\hline
\end{tabular}
\end{adjustbox}
\end{table*}

\section*{Conclusion}
In this paper, we analyzed how age influences observers' gaze distribution during scene exploration using two computational approaches. First, we addressed the explorativeness of an observer which was quantified using the entropy of a saliency map. Our results showed that scene explorativeness increases with age. Second, we measured the average agreement score of the human saliency map of an age group and compared it with other observers of the same or different age groups. This was done by using AUC metrics. In intra-age group prediction analysis, four-year-olds were found to have the highest agreement scores whereas the adult group had the lowest. In inter-age group prediction analysis, we found that an observer from a certain age group better predicted the saliency map of an observer from the same age group. Finally, we proposed an age-adapted framework based on our data analysis for an upgraded version of existing saliency models. We proposed a multi scale feature subset selection from center-surrounded feature maps for different age groups and then learned optimal weights over it. Finally we verified our model for patch-based saliency prediction, which outperform the existing methods of saliency prediction.  
  
\ifCLASSOPTIONcaptionsoff
  \newpage
\fi



%

%


\end{document}